\documentclass[final]{cvpr}

\usepackage{times}
\usepackage{epsfig}
\usepackage{graphicx}
\usepackage{amsmath}
\usepackage{amssymb}
\usepackage{multirow}
\usepackage[ruled]{algorithm2e}

\usepackage[pagebackref=true,breaklinks=true,colorlinks,bookmarks=false]{hyperref}

\def\cvprPaperID{****} 
\def\confYear{CVPR 2021}

\begin{document}

\title{City-Scale Multi-Camera Vehicle Tracking Guided by Crossroad Zones}

\author{
   Chong Liu \textsuperscript{1,2,3}\thanks{The work was done when Chong Liu was intern at Alibaba Group} \and
   Yuqi Zhang \textsuperscript{3} \thanks{Equal contribution} \and
   Hao Luo \textsuperscript{3}\and
   Jiasheng Tang \textsuperscript{3}\and
   Weihua Chen \textsuperscript{3}\and
   Xianzhe Xu \textsuperscript{3}\and
   Fan Wang \textsuperscript{3}\and
   Hao Li \textsuperscript{3}\and
   Yi-Dong Shen \textsuperscript{1} \and
\textsuperscript{1} State Key Laboratory of Computer Science, Institute of Software, Chinese Academy of Sciences, China\\
\textsuperscript{2} University of Chinese Academy of Sciences, Beijing 100049, China \\
\textsuperscript{3} Machine Intelligence Technology Lab, Alibaba Group \\
{\tt\small
\{liuchong,ydshen\}@ios.ac.cn}\\
{\tt\small
\{gongyou.zyq,michuan.lh,jiasheng.tjs,kugang.cwh,xianzhe.xxz,fan.w,lihao.lh\}@alibaba-inc.com}
}

\maketitle

\begin{abstract}
   Multi-Target Multi-Camera Tracking has a wide range of applications and is the basis for many advanced inferences and predictions. This paper describes our solution to the Track 3 multi-camera vehicle tracking task in 2021 AI City Challenge (AICITY21). This paper proposes a multi-target multi-camera vehicle tracking framework guided by the crossroad zones. The framework includes: (1) Use mature detection and vehicle re-identification models to extract targets and appearance features. (2) Use modified JDETracker (without detection module) to track single-camera vehicles and generate single-camera tracklets. (3) According to the characteristics of the crossroad, the Tracklet Filter Strategy and the Direction Based Temporal Mask are proposed. (4) Propose Sub-clustering in Adjacent Cameras for multi-camera tracklets matching. Through the above techniques, our method obtained an IDF1 score of 0.8095, ranking first on the leaderboard \footnote{https://www.aicitychallenge.org/}. The code have released: https://github.com/LCFractal/AIC21-MTMC.
\end{abstract}

\section{Introduction}

The demand for Multi-Target Multi-Camera Tracking (MTMCT) has attracted great attention in these years. Applications such as vehicle tracking helps modern traffic flow prediction and analysis. MTMCT is often split into several sub-tasks: single camera tracking (SCT) and appearance-based feature re-identification (ReID) and trajectory clustering: (1) Single camera tracking is also known as Multiple Object Tracking (MOT) in the community which often follows a tracking-by-detection manner. (2) Re-identification tries to retrieve exactly the same instance from a large gallery set. (3) Trajectory clustering aims to merge the tracklets in cameras into cross-camera links. Although well studied in separate tasks like object detection, tracking and re-identification, an optimized multiple camera multi tracking framework is still promising.

\begin{figure}[t]
   \begin{center}
      \includegraphics[width=0.97\linewidth]{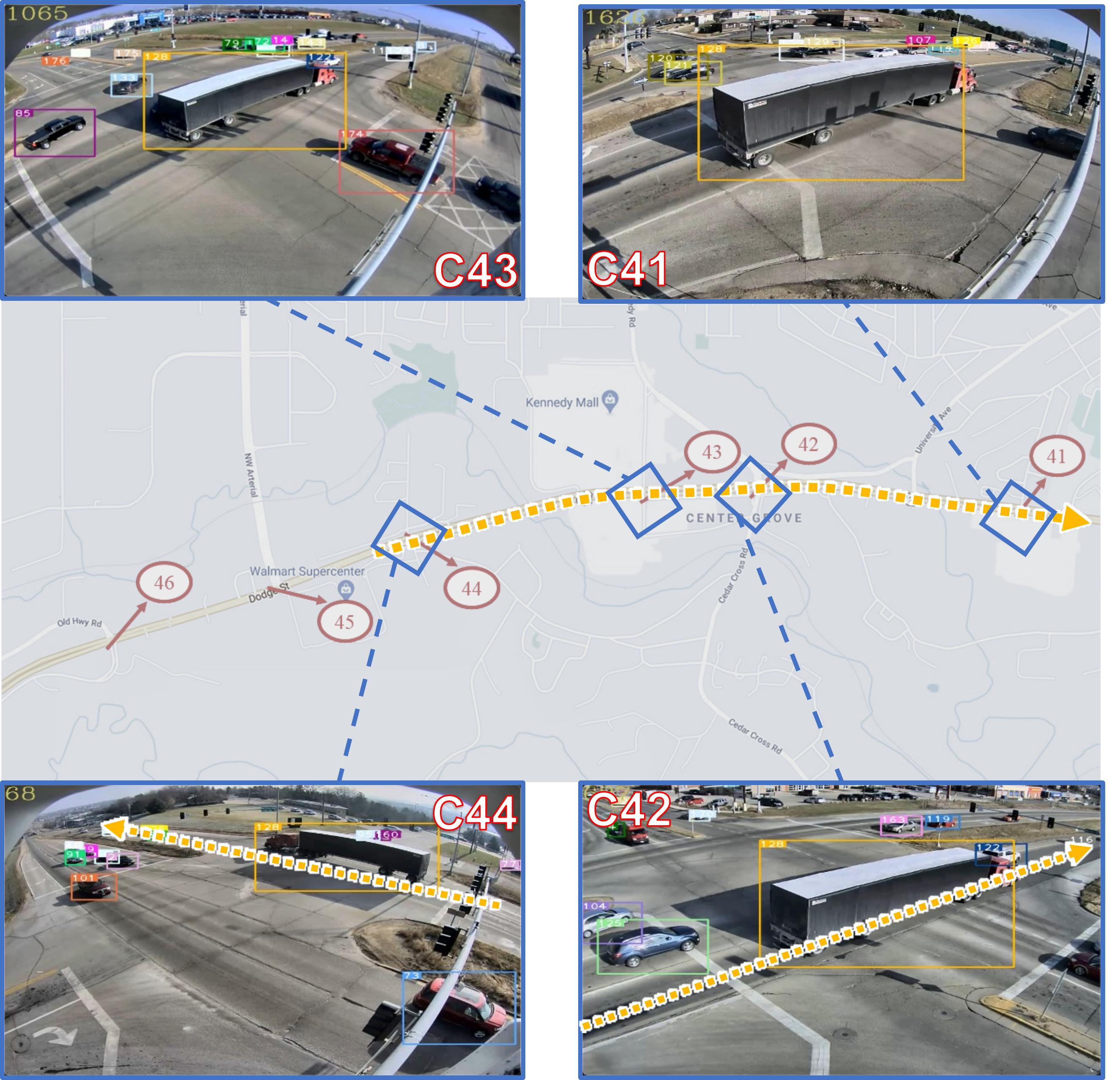}
   \end{center}
   \vspace{-0.8 em}
      \caption{\textbf{Multi-camera vehicle tracking.} Multi-camera multi-target vehicle tracking requires finding the same vehicle that appears in multiple cameras. Their appearance and size usually vary greatly depending on the angle of view and the distance from the camera.}
      \vspace{-1 em}
   \label{fig:introduction}
\end{figure}

For vehicle MTMCT, we observe several challenges: (1) Vehicles are often missed due to distortion or lighting conditions. In the scenario of multiple object tracking, this is often eased by better detectors, better data association strategies or even single object tracking. However, these methods rely on heavy training data or external models. (2) Vehicles share similar appearances and thus pure re-identification models fail to separate these vehicles. (3) The same vehicle instance may suffer from great appearance changes and thus fail to be grouped among different cameras. Considering these challenges and some basic traffic rules, we manage to solve these problems as follows.

For missed vehicles, we propose Tracklet Filter Strategy (TFS). We first set a low threshold for detection and multiple tracking and thus get enough tracklets with a high recall. However, these raw tracklets may contain false positives such as static traffic signs. These static false positives never move the regions and could be filtered. Also as pointed out by~\cite{qian2020electricity}, some vehicles only go through the sub-path without moving into the main road. These tracklets should be filtered to reduce the searching space of the vehicles on the main road.

For vehicles with similar appearance, we propose Direction Based Temporal Mask (DBTM) to constrain the matching space. By setting different zones in each camera, we manage to get the vehicle moving directions. For two tracklets from different cameras, we judge whether they should be disconnected by the corresponding time-stamps and moving directions. For example, a car moving from C42 to C43 at T should not match with any tracklets moving from C42 to C41 since the moving directions conflict. By the simple yet effective temporal mask, the searching space get reduced greatly and thus alleviate the pressure on re-identification. 

For the same vehicle with great appearance changes, we propose Sub-clustering in Adjacent Cameras (SCAC) which tries to match tracklets in adjacent cameras. The motivation is that vehicles always go through continuous cameras, e.g., a vehicle from C41 should not match with C43 without the internal C42. By sub-clustering, the tracklets in adjacent cameras are grouped first and then query expansions are performed on these locally matched tracklets. The query expansion introduces more information for these locally matched tracklets and make them more likely to match with potential tracklets in other cameras.

In summary, we have make the following contributions in this paper:
\begin{itemize}
\item We propose Tracklet Filter Strategy (TFS) to improve precision, which has no demand for any external object detectors.

\item We propose Direction Based Temporal Mask (DBTM) which helps reduce matching space for visual re-identification.

\item We propose Sub-clustering in Adjacent Cameras (SCAC) to merge adjacent tracklets first and then use these matched local tracklets for query expansion. The sub-clustering method helps link the vehicles suffering from great appearance changes.

\end{itemize}

\section{Related Work}

\subsection{Multiple Object tracking}
Currently tracking-by-detection is the dominant scheme for multiple object tracking. Given object detections, multiple object tracking aims to associate them into long tracks by either offline tracking~\cite{tangmin, du20211st} or online tracking~\cite{bewley2016simple, wojke2017simple, zhang2016makes, zhang2017multi}. Offline multi-tracking builds a graph based on visual and spatial-temporal similarities and then optimize the graph for the solution. It often achieves better performance at a cost of more computation time. Online tracking, on the other hand, aims to associate tracks and detections without future information. Some simple yet effective approaches have been proposed including SORT~\cite{bewley2016simple} and Deep-SORT~\cite{wojke2017simple}. With only history information, these methods rely on accurate appearance re-identification model for long-term tracking to deal with occlusions. In recent years, joint detection and tracking~\cite{wang2019towards, zhang2020fair, zhou2020tracking}, single object tracking~\cite{feng2019multi, shuai2020multi} have also been proposed.

\subsection{Object Re-identification}
Object Re-identification (ReID) aims to retrieve the same instance in different scenes. The well studied person re-identification usually studies loss functions~\cite{chen2017beyond, chen2017multi}, part-based models~\cite{sun2018beyond, wang2018learning, luo2019alignedreid++} or unsupervised/semi supervised learning~\cite{ge2020mutual}. Many useful tricks~\cite{luo2019bag, luo2019strong,liu2020unity} have been proposed to set strong baselines for the field. For the topic of vehicle re-identification, more attention has been received due to the applications in city management and intelligent traffic.  The topic has seen multi domain learning~\cite{he2020multi}, large-scale datasets~\cite{lou2019veri}, synthetic data~\cite{zheng2020going} and so on. With the emergence of transformer-based vision tasks, vehicle re-identification has been greatly improved as in ~\cite{he2021transreid}.

\subsection{Trajectory clustering}
With the above two basic modules, multiple camera multi tracking can be regarded as a trajectory clustering problem. Many previous works follow this scheme for MTMCT. Graph-based methods~\cite{chen2014novel, chen2016equalized} establish a global graph for multiple tracklets in different cameras and optimize for a MTMCT solution. Spatial-temporal constraints and traffic rules ~\cite{hsu2019multi, hsu2020traffic, lee2015combined, tang2019cityflow, tang2018single} have been embedded into the clustering stage. With these constraints, the searching space is reduced greatly and thus vehicle re-identification accuracy improves greatly. With the same camera distribution of test data and training data, methods~\cite{tang2019cityflow, tang2018single, hsu2019multi, hsu2020traffic} learn the transition time distribution for each pair of adjacently connected cameras without the need of hand tuning. For MTMCT in completely different test set without knowing camera distribution, methods~\cite{qian2020electricity} observe some basic rules to constrain the matching field.
\begin{figure*}[t]
   \begin{center}
      \includegraphics[width=1\linewidth]{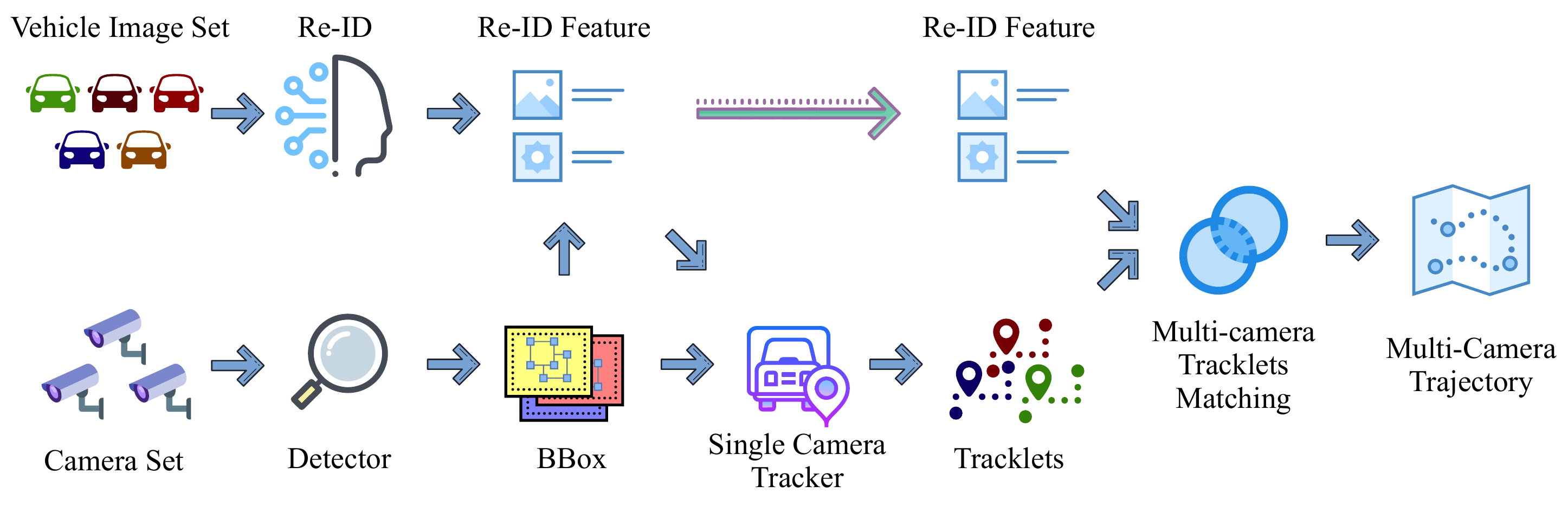}
   \end{center}
      \caption{\textbf{Pipeline of Multi-Target Multi-Camera.} The MTMCT system first uses the detector to obtain the BBox of the target from each frame of each camera video; then uses the trained Re-ID model to extract the apparent features of the target BBox; the single-camera tracker uses BBox and Re-ID feature for each tracking target to generate single-camera tracklets; finally, according to the single-camera tracklets and Re-ID feature, the ID synchronization between cameras generates a multi-camera trajectory.}
   \label{fig:pipeline}
\end{figure*}
\section{Method}

\begin{figure}
   \begin{center}
      \includegraphics[width=1\linewidth]{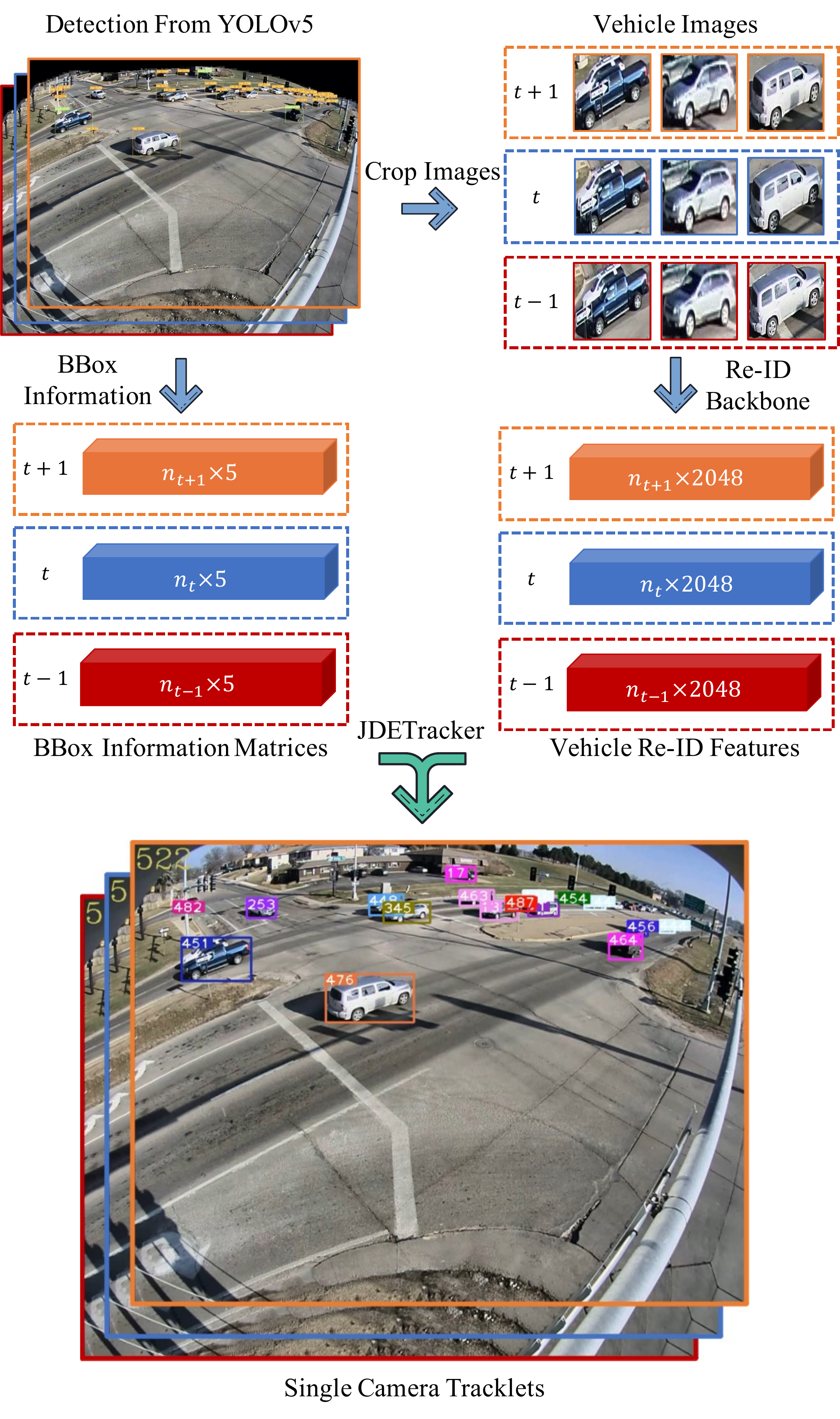}
   \end{center}
      \caption{\textbf{Vehicle single camera tracking framework.} The detector generates vehicle detection for each frame, and uses the Re-ID backbone to generate the corresponding detection appearance features. Modified JDETracker uses BBox information and Re-ID appearance features for single-camera tracking, and generates  single-camera vehicle tracking for each camera.}
   \label{fig:SCT}
\end{figure}

\begin{figure}
   \begin{center}
      \includegraphics[width=1\linewidth]{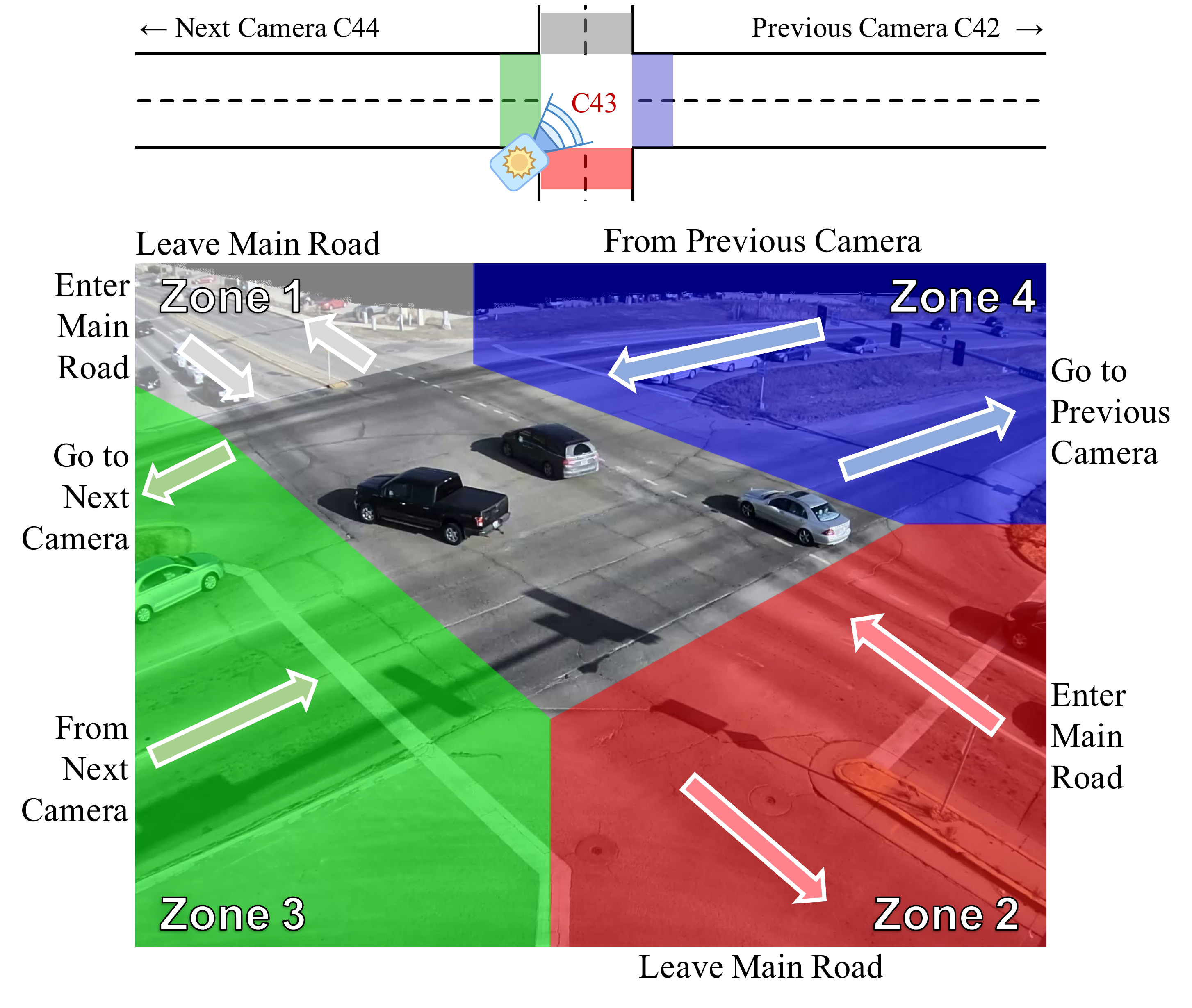}
   \end{center}
      \caption{\textbf{Crossroad zones for C43.} According to the characteristics of the crossroad, we divide the video frame into four zones to divide the vehicle tracklets.}
   \label{fig:ZONE}
\end{figure} 

\subsection{Overview}
The MTMCT system we proposed is shown in Figure \ref{fig:pipeline}. The whole process mainly involves: detection, Re-ID, SCT and MCT. The steps of MTMCT are as follows: 1) Use the detector to obtain the target BBox from each frame of each camera video; 2) Use the trained Re-ID model to extract the Re-ID feature of the target BBox; 3) Single camera tracker uses BBox and Re-ID feature to generate a single-camera tracklets for each tracking target; 4) According to the single-camera tracklets and Re-ID feature, the ID synchronization between cameras generates a multi-camera trajectory. The detailed process will be described below.

\begin{figure*}[t]
   \begin{center}
      \includegraphics[width=1\linewidth]{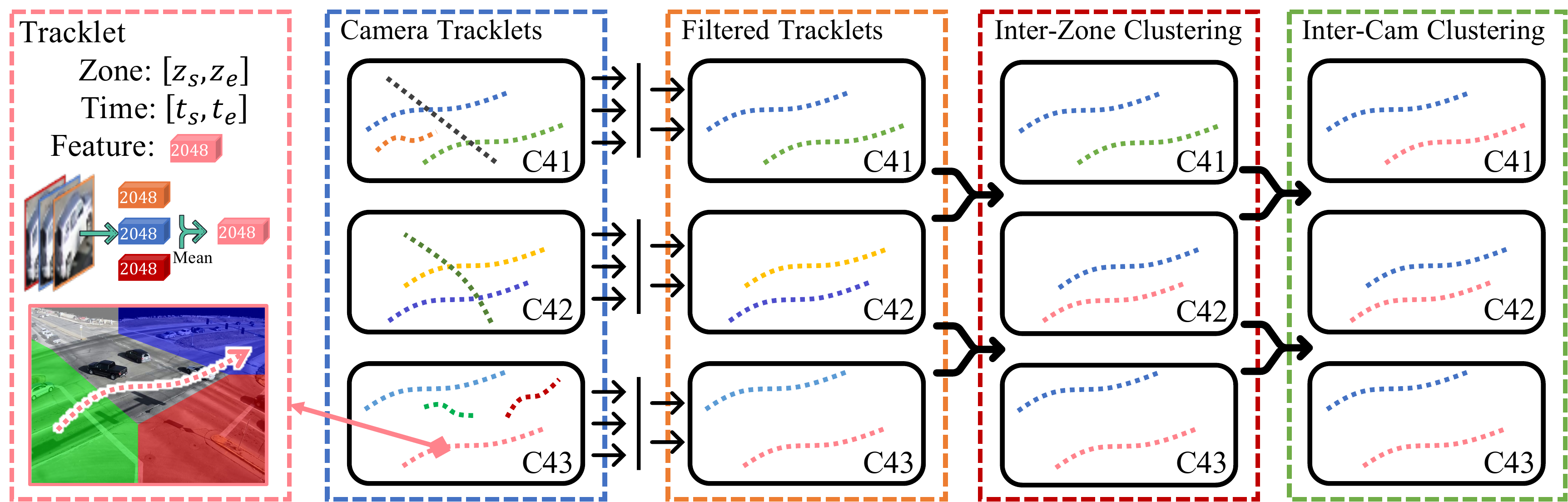}
   \end{center}
      \caption{\textbf{Process of Multi-camera Tracklets Matching.} 1) Generate the necessary information for matching according to crossroad zones and tracklets; 2) Use TFS to filter the tracklets; 3) Calculate the similarity matrix between the tracklets and use DBTM to perform matching constraints; 4) Perform SCAC process between tracklets, perform inter-zone clustering and inter-cam clustering respectively. }
   \label{fig:mcmt}
\end{figure*}

\subsection{Vehicle Detection}
Reliable vehicle detection is a prerequisite for vehicle tracking. We use single-stage YOLOv5~\cite{YOLOv5} with good detection performance to detect vehicles in video frames. We only use the simple and powerful YOLOv5x model pre-trained on the COCO dataset, and do not introduce external data to the detection. With the detection model, we can obtain the BBox of the detected object in each video frame and the corresponding confidence.

The detector will output 80 categories related to the COCO dataset, and there are a large number of categories that are not related to vehicle detection. Therefore, we only test three categories related to vehicles: cars, trucks and buses. At the same time, in order to avoid the same target being detected multiple times by different categories, we perform non-maximum suppression (NMS)~\cite{neubeck2006efficient} for all detected targets. All detection BBoxes in the same video frame are filtered by IoU and confidence score to avoid repeated detections. Finally, we generate a detection BBox for each frame of each camera for subsequent vehicle tracking.

\subsection{Vehicle Re-identification}

For vehicle re-identification, we use the public rei-id strong baseline~\cite{luo2019bag} and train the models with data in Track2. Specifically, the image size is $384 \times 384$ and we set $stride=1$ for the last pooling layer. With these settings, more details of the vehicles could be preserved which helps vehicle re-identification. Different from part-based models, we use the global vehicle feature and perform plenty of data augmentation during training. The re-id network can be trained with loss functions like
\begin{equation}
L_{reid} = L_{cls} + \alpha L_{trp}
\label{eq:qg_rcnn}
\end{equation}
where $L_{cls}$ and $L_{trp}$ stands for softmax cross-entropy loss and triplet loss, with $\alpha$ balancing their weights. The two basic loss functions can be further writen as:
\begin{equation}
L_{cls}=-\log \frac{e^{W_{y_{i}}^{T}} \boldsymbol{x}_{i}+b_{y_{i}}}{\sum_{j=1}^{n} e^{W_{j}^{T} \boldsymbol{x}_{i}+b_{j}}}
\label{eq:softmax}
\end{equation}
where $\boldsymbol{x}_{i} \in \mathbb{R}^{d}$ denotes the $i$-th deep feature, belonging to the $y_{i}$th class. $d$ is the feature dimension. $W_{j} \in \mathbb{R}^{d}$ denotes the $j$th column of the weights $W \in \mathbb{R}^{d \times n}$ in the last fully connected layer and $\boldsymbol{b} \in \mathbb{R}^{n}$ is the bias term.

\begin{equation}
L_{trp}=\left[d_{p}-d_{n}+\alpha\right]_{+}
\label{eq:triplet}
\end{equation}
where $d_{p}$ and $d_{n}$ are feature distances of positive pair and negative pair. $\alpha$ is the margin of triplet loss, and $[z]_{+}$ equals to $max(z, 0)$. 

\subsection{Vehicle Single Camera Tracking}

In single-camera tracking, we associate the detection in the video frame with the corresponding tracklet to achieve single-camera tracking of multi-vehicle targets. FairMOT~\cite{zhang2020fairmot} is the latest single-camera tracking model, which is a unified SCT model for detection and tracking.
We borrow the tracker builder and track management parts, namely Kalman Tracker+Casecade Matching, from JDE and modify them into the vehicle tracking version.
As shown in Figure~\ref{fig:SCT}, we crop the corresponding target image from the detection results, and use the Re-ID model to output the corresponding vehicle Re-ID features. Modified JDETracker uses BBox information matrices and vehicle Re-ID features to assign corresponding tracklet IDs 
with vehicle detection. Finally, the tracker generates a set of triple vectors for each tracklet:

\begin{equation}
   T_{id}=[T_{id,i}:(t_i,b_i,f_i)]
\end{equation}

where, $T_{id}$ is the tracklet corresponding to $id$, $t_i$ is the time frame, $b_i$ is the corresponding BBox information, and $f_i$ is the corresponding Re-ID feature.

\subsection{Multi-camera Tracklets Matching}
Because of the relatively uniform model and color of the vehicle, it is difficult to judge whether the vehicles are the same only from the appearance features. For multi-camera vehicle tracking which involves more matching between similar vehicles, the problem becomes even harder. In this section, we will model the multi-camera vehicle tracking problem at intersections, taking into account the characteristics of the intersection and the space-time constraints between cameras to guide the matching of multi-cameras.

The process of multi-camera tracklets matching is shown in Figure \ref{fig:mcmt}: 1) Generate the necessary information for matching according to crossroad zones and tracklets; 2) Use TFS to filter the tracklets; 3) Calculate the similarity matrix between the tracklets and use DBTM to perform matching constraints; 4) Perform SCAC process between tracklets, perform inter-zone clustering and inter-cam clustering respectively.

\subsubsection{Crossroad Zones}
Taking into account the characteristics of the crossroad, we can easily divide the area in the video. Take C43 as an example (Figure \ref{fig:ZONE}). We use 4 colors to divide different zones (White: Zone 1; Blue: Zone 2; Green: Zone 3; Red: Zone 4). For each camera, two of the zones are connected to the main road, and the vehicle leaves the main road from the other two zones. All cameras are on the main road, so we only need to consider vehicles passing through the main road. Specifically, the roads in zones 1 and 2 cross the main road, and vehicles passing through zones 1 and 2 may enter or leave the main road; zones 3 and 4 can go to other intersections on the main road; zone 3 is connected to the next camera (C44); zone 4 is connected to the previous camera (C43). For each tracklet $T_{id}$, we can get its start-end zone $[z_s,z_e]$ and start-end time $[t_s,t_e]$ under the current camera.
According to crossroad zones, we propose Tracklet Filter Strategy and Direction Based Temporal Mask.\medskip\\
\textbf{Tracklet Filter Strategy (TFS)} is used to filter wrong tracklets from the raw tracklets. These raw tracklets may contain false positives, such as static traffic signs. These static false positives will never move the zone. In addition, some vehicles only pass through sub-paths without entering the main road. TFS filters these trails according to Crossroad Zones to reduce the search space for vehicles on major roads. Specifically, if $z_s=z_e$, it means that the trajectory has not changed in the whole life cycle. We think these are noise or broken trajectories and need to be filtered; if $z_s=1, z_e=2$ or $z_s=2, z_e=1$, then $T_{id}$ does not enter the main road and can not participate in tracklets matching.\medskip\\
\textbf{Direction Based Temporal Mask (DBTM)} is used for matching constraints between tracks. Through crossroad zones, we managed to get the direction of movement.
For two tracklets $T_i$ and $T_j$, we can judge whether they conflict according to the conflict table (Table \ref{tab:conflict}). If they conflict, they cannot match each other. Through DBTM, the search space is greatly reduced, thereby reducing the pressure of re-identification.
Thus, for the two trajectories $T_i$ and $T_j$, we can get the matchability between them as:

\begin{table}
   \small
      \begin{center}
      \begin{tabular}{|l|c|c|c|}
      \hline
      Crossroad Zone & Camera & Time & Cnoflict\\
      \hline\hline
      $z_s^i = 1 \text{ or } 2$ & -         & $t_e^j<t_s^i$ & True \\
      $z_s^i = 3$               & $c^j>c^i$ & $t_e^j>t_s^i$ & True \\
      $z_s^i = 4$               & $c^j<c^i$ & $t_e^j>t_s^i$ & True \\
      \hline\hline
      $z_e^i = 1 \text{ or } 2$ & -         & $t_s^j>t_e^i$ & True \\
      $z_e^i = 3$               & $c^j>c^i$ & $t_s^j<t_e^i$ & True \\
      $z_e^i = 4$               & $c^j<c^i$ & $t_s^j<t_e^i$ & True \\
      \hline
      \end{tabular}
      \end{center}
      \caption{\textbf{Conflict table.}
      When $T_i$ and $T_j$ meet the three conditions of Crossroad Zone, Camera and Time at the same time, $T_i$ and $T_j$ do not match.}
      \label{tab:conflict}
   \end{table}

\begin{equation}
\mathrm{mask}(T_i,T_j)= \left\{\begin{array}{cc}0&\text{cnoflict}\\1&\text{o.w.}\end{array}\right.
\end{equation}

For $m$ tracklets, we can get the DBTM matrix between the trajectories:
\begin{equation}
M=\begin{bmatrix}
   \mathrm{mask}(T_1,T_1)  & \dots & \mathrm{mask}(T_1,T_m) \\
   \vdots & \ddots & \vdots  \\
   \mathrm{mask}(T_m,T_1)  & \dots & \mathrm{mask}(T_m,T_m)
  \end{bmatrix}.
\end{equation}

\subsubsection{Similarity Matrices and Reranking}
Before starting to match, we need to calculate the similarity between each trajectory. All trajectory features are represented by 2048-dimensional averaged features for all frames:
\begin{equation}
   T_{id}=[T_{id,i}:(t_i,b_i,f_i)].
\end{equation}
The appearance similarity of tracklets $T_i$ and $T_j$ can be computed using cosine similarity:
\begin{equation}
   \cos(T_i,T_j)=\frac{F(T_i)\times F(T_j)}{||F(T_j)||\times||F(T_j)||},
\end{equation}
where, $F(T_i)$ is the average feature of trajectory $T_i$, and $F(T_j)$ is the average feature of trajectory $T_j$. From this we can get the similarity matrix $S$ between $m$ trajectories:
\begin{equation}
   S=\begin{bmatrix}
      \cos(T_1,T_1)  & \dots & \cos(T_1,T_m) \\
      \vdots & \ddots & \vdots  \\
      \cos(T_m,T_1)  & \dots & \cos(T_m,T_m)
     \end{bmatrix}.
\end{equation}\\
The similarity matrix might be still imperfect due to severe illumination or view changes. We focus on the camera bias, which influences the discrimination of the model. The mean value of features under the same camera is subtracted from each tracklet feature. Then the tracklet feature is updated with its closest neighbours. We also perform k-reciprocal reranking method~\cite{zhong2017re} to refine the updated simialrity matrix. The k-reciprocal neighbours of the tracklets are enhanced greatly and thus generates a stronger similarity matrix $S$. 
Finally, we combine the similarity matrix with DBTM to obtain the DBTM similarity matrix $\hat{S}$:
\begin{equation}
   \hat{S}=S\odot M,
\end{equation}
where, $\odot$ represents the product of the corresponding elements of the matrix. The constrained similarity matrix is thus obtained, which is used for subsequent tracklets clustering.

\subsubsection{Sub-clustering in Adjacent Cameras}
For the tracklets matching between cameras, the commonly used method is to perform hierarchical clustering of all trajectories according to the DBTM similarity matrix $\hat{S}$. This kind of method performs clustering in a huge range with all cameras, and it is difficult to gather the correct vehicles together. At the same time, it may gather with the wrong vehicles and lead to wrong clusters. According to the characteristics of the data set scene, we proposed Sub-Clustering in Adjacent Cameras (SCAC). 
This is a local clustering method based on hierarchical clustering. As shown in Figure \ref{fig:mcmt}, it is mainly divided into two processes: inter-zone clustering and inter-cam clustering.\medskip\\
\textbf{Inter-zone clustering} is used to cluster between the zones of different cameras. In the data set, the current camera zone 4 is connected to the zone 3 of the previous camera, and the current camera zone 3 is connected to the zone 4 of the remaining camera. Therefore, we first perform hierarchical clustering on the trajectories in the connected zones to ensure high-confidence vehicle clustering and ensure the correctness of the clustering.\medskip\\
\textbf{Inter-cam clustering} is used for clustering between connected cameras. It is used to cluster all tracklets in the camera on the basis of inter-zone clustering to ensure that trajectories that cannot be described by Crossroad Zones can be matched. The broadness of the class.\medskip\\
Through these two clustering methods, we can match as many trajectories as possible while still ensuring accuracy. The final inter-camera tracklets matching result is then obtained and merged into a complete trajectory.

\section{Experiment}
\begin{table}
   \small
      \begin{center}
      \begin{tabular}{|c|c|c|c|}
      \hline
      Rank & Team ID & Team Name & IDF1 Score\\
      \hline\hline
      \textbf{1} & \textbf{75} & \textbf{mcmt (Ours)} & \textbf{0.8095} \\
      2  & 29  & fivefive         & 0.7787 \\
      3  & 7   & CyberHu        & 0.7651 \\
      4  & 85  & FraunhoferIOSB & 0.6910 \\
      5  & 42  & DAMO           & 0.6238 \\
      6  & 27  & Janus Wars     & 0.5763 \\
      7  & 15  & aiforward      & 0.5654 \\
      8  & 48  & BUPT-MCPRL2    & 0.5534 \\
      9  & 79  & oOIAMAIOo      & 0.5458 \\
      10 & 112 & Dukbaegi       & 0.5452 \\
      \hline
      \end{tabular}
      \end{center}
      \caption{\textbf{Leaderboard of City-Scale Multi-Camera Vehicle Tracking.} Our method takes the first place in 2021 AI City Challenge Track 3.}
      \label{tab:results}
   \end{table}
\subsection{Dataset and Evaluation Setting}
\subsubsection{Dataset}
This paper uses the CityFlow~\cite{tang2019cityflow} dataset for evaluation. CityFlow is the largest and most representative MTMCT data set captured in the actual scene of the city. In the training set and validation set, it contains 3.25 hours of traffic video of 40 cameras at 10 intersections in a medium-sized city, with a total length of about 2.5 kilometers. In addition, CityFlow covers a variety of different road traffic types, including intersections, road extensions and highways. For the test set, it contains 6 intersections for the competition.

\begin{figure*}[h]
   \begin{center}
      \includegraphics[width=1\linewidth]{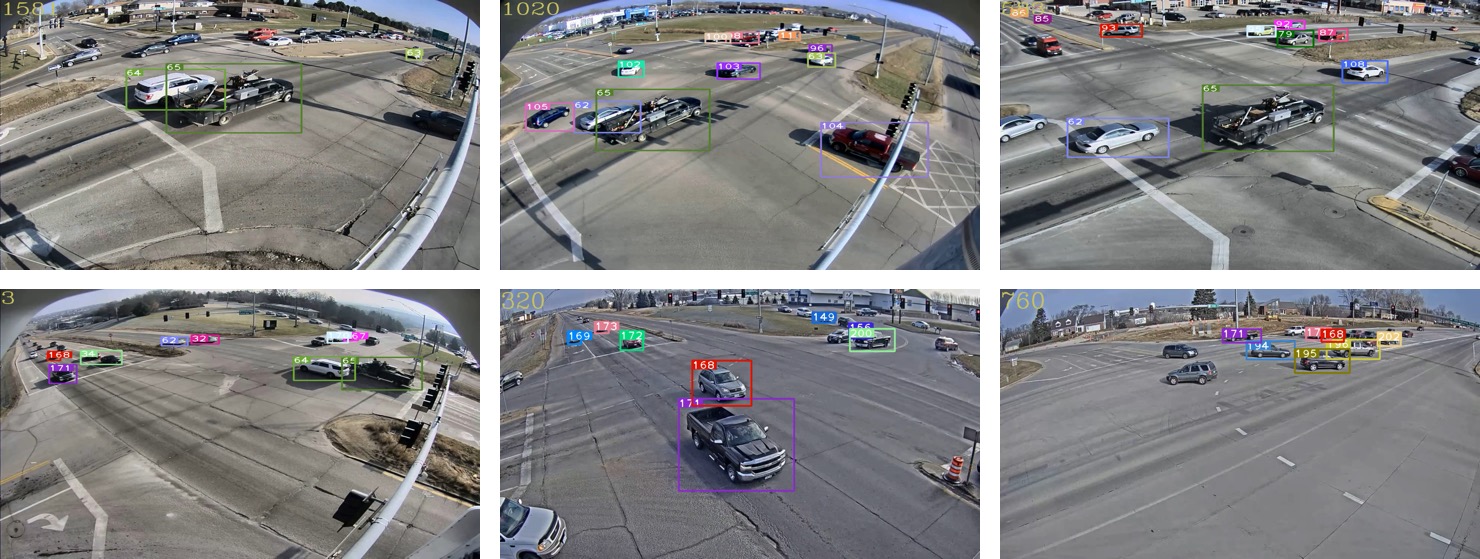}
   \end{center}
   \vspace{-0.5 em}
      \caption{\textbf{Visualization of vehicle multi-camera tracking results.} }
   \label{fig:viz}
\end{figure*}

\subsubsection{Evaluation Metrics}
For MTMCT, we use IDF1, IDP and IDR as evaluation indicators. IDF1~\cite{Duck} calculates the ratio of the number of correctly identified detections to the ground truth and the average number of calculated detections. More specifically, the false negative ID (IDFN), true negative ID (IDTN) and true positive ID (IDTP) counts are all used to calculate the  IDF1 score: 
\begin{align}
   \begin{split}
   IDF1&=\frac{2IDTP}{2IDTP+IDFP+IDFN}.
   \end{split}
\end{align}

\subsection{Implementation Details}
Our algorithm is implemented in PyTorch 1.7.1 and is performed on eight Tesla P100 GPUs. In the vehicle Re-ID training process, we respectively use ResNet50-IBN-a~\cite{pan2018two}, ResNet101-IBN-a~\cite{pan2018two} and ResNeXt101-IBN-a~\cite{xie2017aggregated} as the backbone for training and inference. In the detection process, we use the YOLOv5x model pre-trained on COCO to perform vehicle detection with a confidence threshold of 0.1. In the SCT process, we use modified JDETracker to perform single-camera vehicle tracking with a confidence threshold of 0.1 and an area threshold of 750 pixels. In the MCT process, we performed SCAC camera tracklets matching with a distance threshold of 0.2, and finally generated 226 cross-camera trajectories. Our method takes the first place in 2021 AI City Challenge Track 3 with IDF1 0.8095.

\subsection{Ablation Study}
\begin{table}
   \small
      \begin{center}
         \begin{tabular}{|l|c|c|c|c|c|}
         \hline
         Method & IDF1 & IDP & IDR & Precision & Recall \\
         \hline\hline
         Baseline & 30.48 & 22.42 & 47.61 & 30.94 & 65.71 \\
         +TFS     & 34.88 & 32.84 & 37.20 & 42.76 & 48.44 \\ 
         +DBTM    & 53.94 & 57.38 & 50.89 & 67.17 & 59.56 \\
         +Rerank  & 63.61 & 69.53 & 58.61 & 78.32 & 66.01 \\
         +SCAC    & 67.77 & 77.57 & 60.16 & 85.32 & 66.18 \\
         \hline
         \end{tabular}
         \end{center}
         \caption{\textbf{The performance of each module on the leaderboard.} With the increase of modules, the performance of the model is better and better}
         \label{tab:ablation}
   \end{table}

\begin{table}
   \small
      \begin{center}
         \begin{tabular}{|l|c|c|c|c|c|}
         \hline
         Backbone & IDF1 & IDP & IDR & Precision & Recall \\
         \hline\hline
         IBNR50 - & 67.77 & 77.57 & 60.16 & 85.32 & 66.18 \\
         IBNR101& 78.46 & 82.06 & 75.16 & 85.44 & 78.26 \\
         IBNR101*& 79.81 & 85.06 & 75.18 & 87.90 & 77.69 \\
         IBNRX101*& 79.82 & 85.72 & 74.68 & 88.66 & 77.24 \\
         \textbf{Merge*}& \textbf{80.95} & \textbf{85.69} & \textbf{76.70} & \textbf{88.14} & \textbf{78.90} \\
         \hline
         \end{tabular}
         \end{center}
         \caption{\textbf{The performance of each backbone on the leaderboard.} - means that the model hyperparameters are not adjusted. * means flip.}
         \label{tab:reid}
\end{table}
Table~\ref{tab:ablation} shows the effect of using each module separately on the results. We verified the effects of the proposed TFS, DBTM, Rerank and SCAC on performance. Our baseline uses ResNet50-IBN-a as the backbone. TFS reduces the recall to a certain extent by filtering the tracklets, but greatly improves the Precision, and provides a reference for subsequent modules. DBTM, Rerank and SCAC respectively further improve the performance of the model.

In addition, Table \ref{tab:reid} verifies the influence of different backbone networks (ResNet50-IBN-a, ResNet101-IBN-a and ResNeXt101-IBN-a) on the model. Finally, the best performance was achieved by merge ResNet101-IBN-a and ResNeXt101-IBN-a.

\section{Conclusion}

This paper proposes a multi-camera vehicle tracking framework guided by crossroad zones. Based on the mature tasks of detection, Re-id, and single-camera tracking, we propose an crossroad zone method for multi-camera vehicle tracking. According to the crossroad zone, we proposed three modules, TFS, DBTM, and SCAC to improve the performance of tracking tasks. Our ablation analysis verified the effectiveness of three modules. Our method obtained an IDF1 score of 0.8095, ranking first on the leaderboard

\section*{Acknowledgments}
This work is supported in part by China National 973 program 2014CB340301.

{\small
\bibliographystyle{ieee_fullname}
\bibliography{egbib}
}

\end{document}